**Enhancing Collective Intelligence in Large Language Models Through Emotional Integration**

**Likith Kadiyala**[1,2] (likithanoopvenkata-kadiyala@uiowa.edu)
**Ramteja Sajja**[1,3] (ramteja-sajja@uiowa.edu)
**Yusuf Sermet**[1] (msermet@uiowa.edu)
**Ibrahim Demir**[4,5] (idemir@tulane.edu)

[1] IIHR - Hydroscience and Engineering, University of Iowa, Iowa City, Iowa, U.S.A
[2] Computer Science, University of Iowa, Iowa City, Iowa, U.S.A
[3] Electrical and Computer Engineering, University of Iowa, Iowa City, Iowa, U.S.A
[4] River-Coastal Science and Engineering, Tulane University, New Orleans, Louisiana, U.S.A
[5] ByWater Institute, Tulane University, New Orleans, Louisiana, U.S.A

Corresponding Author Name, Email: Likith Kadiyala, likithanoopvenkata-kadiyala@uiowa.edu

**Abstract**
This research investigates the integration of emotional diversity into Large Language Models (LLMs) to enhance collective intelligence. Inspired by the human wisdom of crowds phenomenon, where group decisions often outperform individual judgments, we fine-tuned the DarkIdol-Llama-3.1-8B model using Google's GoEmotions dataset and Low-Rank Adaptation (LoRA) to simulate emotionally diverse responses. Evaluating the model on a distance estimation task between Fargo, ND, and Seattle, WA, across 15,064 unique persona configurations, we analyzed how emotional states and social attributes influence decision-making. Our findings demonstrate that emotional integration shapes response patterns while maintaining acceptable prediction accuracy, revealing its potential to enhance artificial collective intelligence. This study provides valuable insights into the interplay of emotional diversity and decision-making in LLMs, suggesting pathways for creating emotionally aware AI systems that balance emotional depth with analytical precision.

## 1. Introduction

The concept of collective intelligence, first observed in natural phenomena and human group behaviors, has become increasingly relevant in the era of artificial intelligence (Hosseini et al., 2021). The wisdom of crowds phenomenon, where aggregated group decisions often outperform individual judgments (Prelec et al., 2017; Rosenberg et al., 2016), has traditionally been studied in human decision-making contexts. Research has demonstrated that different approaches to collective decision-making, such as crowd-based aggregation versus swarm-based real-time systems, can significantly impact the quality of group decisions (Rosenberg et al., 2016). Findings further suggest that diversity within intelligent collectives reduces susceptibility to premature consensus, fostering improved deliberation and collective performance (Nguyen et al., 2018).

Recent studies have also quantified the diversity of group members through semantic analysis of social media communications (Sit et al., 2020), demonstrating that diverse virtual crowds can outperform non-diverse or randomly sampled groups, further emphasizing the importance of diversity in collective intelligence (Bhatt et al., 2017). However, its application to artificial intelligence systems, particularly Large Language Models (LLMs), presents both unique opportunities and challenges (GeeksforGeeks, 2023; AI Wiki, 2023; Vald et al., 2024).

Recent advances in LLMs have demonstrated remarkable capabilities in processing and generating human-like text (Dou et al., 2022; OpenAI, 2024; Samuel et al., 2024). Simultaneously, research in emotional intelligence and affective computing has highlighted the crucial role of emotional understanding in decision-making processes (Hume AI, 2024). Emotionally intelligent AI agents have been shown to benefit significantly from cognitive architectures that integrate emotional processing, drawing inspiration from cognitive psychology and neuroscience (Grig & Rizzo, 2023). AI applications in hydrology and hazard mitigation have demonstrated the effectiveness of integrating multimodal LLMs (Pursnani et al., 2024) and structured multiagent frameworks to improve community resilience and disaster risk reduction (Kadiyala et al., 2024a; 2024b). This intersection presents an intriguing opportunity to enhance artificial collective intelligence through emotional integration (Suleyman, 2024; Dubois et al., 2024).

Traditional implementations of LLMs typically generate responses from a singular, emotionally neutral perspective (Wang, X. et al., 2023; Huang & Siddarth, 2024). This approach, while effective for many tasks, may not fully capture the nuanced dynamics of human collective decision-making, where emotional diversity often contributes to more robust outcomes. LLMs handle emotional prompts by leveraging their ability to recognize and generate text that reflects human-like emotions. Recent advances have shown promising developments in emotional support capabilities, with larger models being used to enhance the performance of smaller, specialized emotional support systems (Zheng et al., 2024).

The integration of cultural awareness in emotion recognition systems has emerged as a crucial factor in improving the robustness and accuracy of emotional understanding across diverse cultural contexts (Mehrdad et al., 2024). However, their responses often lack emotional depth and consistency due to several inherent limitations. These limitations stem from the models' dependency on annotated data, the complexity of emotion processing, and the influence of societal and cultural biases. While models like GPT-4 have achieved near-human performance in identifying and describing emotions (Patel & Fan, 2024), they often produce emotionally shallow responses that fail to exhibit nuanced emotional understanding (Chen & Xiao, 2024).

The emotional responses generated by LLMs are further constrained by the training data, which may not fully represent the diversity of human emotional experiences and cultural interpretations of emotions (Chen & Xiao, 2024). Additionally, inconsistencies arise in emotional outputs due to conflicting cues present in diverse training datasets (Li et al., 2023). Gender biases in emotion attribution are also prevalent, with LLMs often reinforcing societal stereotypes, such as associating anger more with men and empathy with women (Plaza-del-Arco et al., 2024). Emerging techniques, such as the Emotional chain of thought (ECoT), have been proposed to align emotional outputs more closely with human emotional intelligence guidelines (Li et al., 2023). Future directions may explore unsupervised learning approaches and interpretable emotion cognition models to address these limitations (Chen & Xiao, 2024).

The relationship between emotional intelligence and decision accuracy in LLMs presents a complex challenge. While some research suggests that emotional awareness can enhance model performance (Zhao et al., 2024), others highlight potential trade-offs between emotional sophistication and analytical precision (Kang et al., 2024). This tension becomes particularly relevant when attempting to simulate crowd wisdom effects in artificial systems (ScienceDaily, 2017). Theories from psychology and behavioral science, such as emotional contagion and affective forecasting, are intricately linked to the concept of collective intelligence in AI. These theories provide a foundation for understanding how individual psychological processes contribute to the emergence of collective intelligence in human-AI systems.

Concepts such as the *Theory of Mind* offer a cognitive framework for modeling the mental states of teammates, which can improve communication, prediction, and team performance in AI systems (Harré et al., 2024; Westby & Riedl, 2022). Bayesian agents, leveraging generative computational approaches, have demonstrated an ability to enhance collective intelligence by predicting and adapting to team dynamics based on observed communication patterns (Westby & Riedl, 2022). The integration of psychological insights, particularly from the *Theory of Mind*, enhances the ability of AI systems to model and predict human behavior, thereby improving the collective intelligence of human-AI teams. This integration is critical for developing sociotechnical systems that effectively combine human and AI capabilities.

The development of frameworks like *Collective Human-Machine Intelligence (COHUMAIN)* emphasizes the importance of cognitive architectures that facilitate seamless human-AI collaboration (Gupta et al., 2023). These frameworks align closely with the *transactive systems model of collective intelligence*, which highlights the critical cognitive processes underlying human-AI teamwork (Gupta et al., 2023). Recent discourse has highlighted the complex interplay between collective intelligence and social contagion in decision-making processes (Arthur, 2024; Huang & Siddarth, 2024).

Group dynamics significantly influence individual perception and judgment, potentially leading to compromised decision-making when collective experiences override personal discernment. This phenomenon raises important considerations for AI systems designed to leverage collective intelligence, as they must balance the benefits of aggregated wisdom against the risks of amplifying groupthink or social contagion effects. The challenge becomes particularly relevant when developing emotionally aware AI systems that interact with human groups, where the potential for emotional contagion must be carefully managed to maintain the integrity of collective decision-making processes.

However, humans often struggle to integrate information from teammates, especially under high communication loads, and exhibit cognitive biases that affect decision-making. AI systems incorporating Theory of Mind principles can help mitigate these limitations by providing real-time insights into team dynamics and improving overall team performance (Westby & Riedl, 2022; Plaza-del-Arco et al., 2024). Despite these advancements, challenges remain in balancing emotional sophistication with analytical precision, addressing biases, and maintaining consistency across emotionally charged contexts.

While the integration of psychological theories into AI systems holds promise for enhancing collective intelligence, it also raises important ethical considerations. For instance, the potential for AI to subtly influence human emotional states and decision-making raises concerns about autonomy, transparency, and fairness in human-AI collaborations. These challenges must be addressed to ensure that emotionally aware AI systems align with human values and foster trust.

Our research addresses these critical gaps by investigating how emotional integration affects the collective intelligence capabilities of LLMs. We utilize the comprehensive *GoEmotions* dataset (Demszky et al., 2020) to fine-tune a *DarkIdol-Llama* model (QuantFactory, 2024), incorporating a range of emotional states while maintaining the model's fundamental predictive capabilities. This approach allows us to examine whether emotional diversity in artificial systems can parallel the benefits observed in human crowd wisdom (Mohammad & Kiritchenko, 2018; Wang, X. et al., 2023). The study makes significant contributions to the field by introducing a novel methodology for incorporating emotional diversity into LLM-based collective intelligence systems. We provide empirical evidence on the relationship between emotional context and prediction accuracy while developing a framework for balancing emotional awareness with analytical precision in artificial systems (Creangă & Dinu, 2024). Our research offers valuable insights into optimizing group size for emotionally enhanced collective intelligence.

Through this study, we aim to bridge the gap between traditional crowd wisdom theory and modern AI capabilities (Hu et al., 2021), advancing our understanding of how emotional contexts influence machine learning-based collective intelligence. These findings have important implications for developing more sophisticated AI systems that can better leverage the benefits of diverse perspectives, whether emotional or analytical (Wang, X. et al., 2023). Our work builds upon recent developments in emotional AI while addressing the largely unexplored question of how emotional diversity affects collective intelligence in artificial systems. By examining this intersection, we contribute to both the theoretical understanding of artificial collective intelligence and its practical implementation in emotionally aware systems (Zhao et al., 2024; Yang et al., 2024).

The remainder of this paper is organized as follows: Section 2 reviews related work on collective intelligence, emotional integration in artificial systems, and fine-tuning techniques for LLMs. Section 3 details our methodology, including system architecture, data preparation, fine-tuning strategy, and experimental setup. Section 4 presents our results and discusses the impact of emotional integration on model performance, followed by Section 5, which concludes with key findings, implications, and future research directions.

## 1.1. Related Work

The intersection of crowd wisdom and artificial intelligence has emerged as a compelling research domain. Previous studies exploring crowd wisdom simulation through LLMs have significantly advanced our understanding of collective intelligence in artificial systems. Research by (Dou et al., 2022) developed the SCARECROW framework for analyzing GPT-3 outputs, demonstrating how multiple model instances could generate diverse, human-like responses. Their work revealed the potential for LLMs to simulate crowd behavior, though primarily focusing on textual characteristics rather than emotional aspects.

In the educational domain, AI-enabled intelligent assistants and tools such as the Educational AI Hub have demonstrated their ability to enhance personalized and adaptive learning experiences by reducing cognitive load and tailoring resources to individual needs (Sajja et al., 2024a; 2024c). Similarly, frameworks for vocational training, such as AI-assisted floodplain manager certification preparation, illustrate the transformative potential of AI in professional education (Sajja et al., 2024b).

Natural Language Processing (NLP) serves as a versatile tool across various disciplines, extending its transformative potential beyond education and vocational training into areas such as health-related studies (Sermet and Demir, 2021; Zhang et al., 2023). The integration of emotions into NLP has seen substantial evolution in recent years. (Mohammad & Kiritchenko, 2018) established foundational work through their comprehensive study of affect interactions in social media data, developing methods to understand complex emotional expressions in text. Their research highlighted the importance of considering emotional context in natural language understanding tasks. Building on this foundation, (Wang, Y. et al., 2023) demonstrated the effectiveness of combining LLMs with emotional intelligence through their work on empathy detection and emotion classification, achieving significant improvements in emotional understanding tasks.

The GoEmotions dataset, introduced by (Demszky et al., 2020), represents a breakthrough in emotion recognition research. This dataset provides unprecedented granularity in emotional categorization, containing 58,000 carefully annotated Reddit comments labeled with 27 distinct emotions. Unlike previous emotion datasets that focused on basic emotional states, GoEmotions captures subtle emotional nuances and mixed emotional states. The dataset's origin in social media conversations provides natural examples of emotional expression in modern communication contexts, making it particularly valuable for training AI systems to understand the complexity of human emotional expression.

Low-Rank Adaptation (LoRA) has revolutionized the fine-tuning of LLMs. The technique, introduced by (Hu et al., 2021), addresses the computational challenges of fine-tuning billion-parameter models by updating only a small number of adaptable parameters while keeping most of the model frozen. Similarly, methods combining model compression and transferable prompts have been introduced to optimize the trade-off between accuracy and efficiency in LLM inference, enabling resource-efficient deployment on constrained hardware (Xu et al., 2023). Recent advancements in LoRA include Bayesian reparameterization techniques to enhance stability and performance during fine-tuning, addressing sensitivity in hyperparameter selection, and improving efficiency in task-specific adaptations (Sengupta et al., 2024).

Building upon these approaches, IVON-LoRA, a variational Bayesian method, introduces a variational objective that improves accuracy and calibration without significantly increasing computational costs, providing an efficient alternative for model adaptation (Bai et al., 2024). These advancements make it possible to efficiently adapt large models for specific tasks while maintaining their general capabilities. The effectiveness of LoRA in emotion-specific tasks has been demonstrated through various implementations, significantly reducing computational requirements while maintaining performance comparable to full fine-tuning approaches.

The convergence of these research streams - crowd wisdom simulation, emotion integration in NLP, comprehensive emotion datasets, efficient fine-tuning techniques, and domain-specific AI tools - creates new opportunities for enhancing LLMs' collective intelligence capabilities. Our work builds upon these foundations while addressing the previously unexplored intersection of emotional diversity and crowd wisdom in artificial intelligence systems.

## 2. Methodology

This study employs a systematic methodology to explore the integration of emotional diversity into LLMs and its impact on collective intelligence. The framework includes data processing, model fine-tuning, and evaluation, leveraging the DarkIdol-Llama-3.1-8B model, the GoEmotions dataset, and unique persona configurations to analyze the interplay between emotional states, social attributes, and model performance.

### 2.1. System Architecture

The experimental framework consists of three main components: data processing, model fine-tuning, and evaluation systems (Figure 1). At its core, the architecture leverages the DarkIdol-Llama-3.1-8B model (QuantFactory, 2024) as the foundation, enhanced through emotional integration using the GoEmotions dataset (Demszky et al., 2020). The data processing pipeline handles two parallel streams: emotional data preparation and social attribute integration. The emotional stream processes the GoEmotions dataset, converting Reddit comments and their associated emotion labels into a format suitable for model fine-tuning. Simultaneously, the social attribute system manages the generation and validation of 15,064 unique persona configurations.

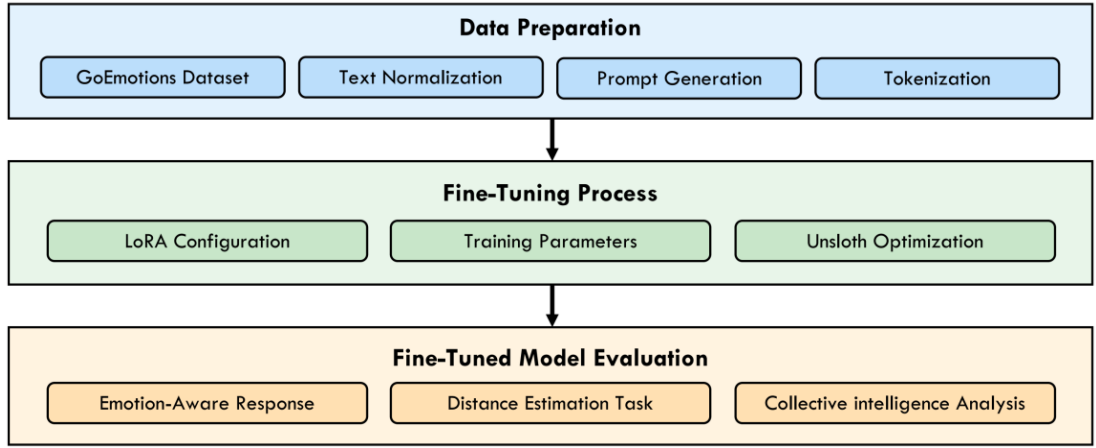

Figure 1. System architecture for emotion-integrated collective intelligence in LLMs

The fine-tuning system employs LoRA (Hu et al., 2021) for efficient model adaptation, implemented through the Unsloth framework (Unsloth, 2024). This component manages the integration of emotional understanding into the base model while maintaining computational efficiency through 4-bit quantization and optimized memory management. The evaluation system orchestrates the testing process by: (i) generating diverse prompts combining emotional states and social attributes; (ii) managing response collection and aggregation; (iii) calculating accuracy metrics within the specified range (1,411-1,441 miles); and (iv) determining optimal subset sizes for maximum accuracy.

## 2.2. Model Selection

For this research, we selected the DarkIdol-Llama-3.1-8B model as our base architecture. This model represents a significant advancement in the Llama model family, offering a balance between computational efficiency and performance capability. Built on Meta's Llama architecture, it incorporates 8 billion parameters, positioning it in the medium-size range of LLMs while maintaining robust performance characteristics.

DarkIdol-Llama-3.1-8B is particularly suited for our research objectives for several key reasons. First, its architecture inherits the efficient scaling properties of the Llama family, allowing for effective fine-tuning without excessive computational demands. The model's 8 billion parameter size provides sufficient complexity to capture nuanced emotional expressions while remaining manageable for experimental manipulation through LoRA fine-tuning techniques.

Another crucial factor in our selection was the model's instruction-following capabilities. The "Instruct" designation in the model's name indicates its optimization for following specific instructions, which is essential for our experimental design where we need precise control over the model's responses to emotional prompts. This characteristic enables us to maintain consistency in our testing framework while introducing emotional variations.

The model's uncensored nature also played a role in our selection. While maintaining ethical guidelines, this characteristic allows for a broader range of emotional expressions, which is particularly important when working with the full spectrum of emotions present in the GoEmotions dataset. This feature enables more natural and unrestricted emotional responses, better simulating the diversity of human emotional expressions found in real-world crowd wisdom scenarios. It's worth noting that the model's availability through the Hugging Face platform (QuantFactory, 2024) facilitated straightforward implementation and reproducibility of our research. The platform's infrastructure provided the necessary tools for both model deployment and fine-tuning procedures, streamlining our experimental process.

### 2.3. Data Preparation

Our research utilized two primary data sources: the GoEmotions dataset for emotional context and a structured set of social attributes for persona generation. The GoEmotions dataset (Demszky et al., 2020) underwent systematic preprocessing to prepare it for fine-tuning. Starting with 58,000 Reddit comments labeled across 27 distinct emotions, we implemented text normalization procedures while preserving the emotional indicators crucial for our research. The dataset's emotional granularity, ranging from basic emotions like joy and sadness to more nuanced states such as gratitude and curiosity, provided the foundation for our emotional diversity framework. For social attributes, we developed a comprehensive persona system comprising 15,064 unique configurations. Table 1 presents the complete attribute framework used in our study.

Each attribute category was designed to capture a specific dimension of human diversity. The age range (18-80) ensures representation across different life stages, while the gender options acknowledge contemporary understanding of gender identity. Occupational categories span various sectors, from academic to creative fields, while personality traits and communication styles reflect different approaches to interaction and problem-solving.

The integration of these attributes with emotional states created a multidimensional testing framework. During our experiments, each test instance combined a specific emotional state from the GoEmotions dataset with a unique persona configuration from this attribute framework. This approach allowed us to systematically examine how different combinations of emotions and social characteristics influenced the accuracy of distance estimation tasks, providing a robust foundation for analyzing collective intelligence in our fine-tuned language model.

The curation of these attributes ensured that each persona combination remained realistic and internally consistent, avoiding contradictory or implausible configurations that might skew our results. This attention to detail in data preparation was crucial for maintaining the ecological validity of our research and enabling meaningful insights into the relationship between emotional diversity and crowd wisdom in artificial intelligence systems.

### 2.4. Fine-Tuning Strategy

This fine-tuning approach implemented a structured process to integrate emotional understanding into the DarkIdol-Llama model while maintaining computational efficiency. As illustrated in Figure 2, the strategy comprised three main components: data preprocessing, LoRA configuration, and training configuration. The data preprocessing phase began with the GoEmotions dataset (Demszky et al., 2020), containing 58,000 labeled examples. We implemented a specialized prompt template for emotion-based formatting and applied tokenization with a maximum sequence length of 2048 tokens.

Table 1. Role attributes and their options (Kadiyala et al., 2024b)

| Attribute | Options |
|---|---|
| *Age* | 18 - 80 |
| *Gender* | Nondisclosed, Female, Genderqueer, Male |
| *Occupation* | Student, Retired, Engineer, Unemployed, Teacher, Doctor, Artist, Scientist |
| *Personality Traits* | Extroverted, Traditional, Open to Experience, Pessimistic, Innovative, Introverted |
| *Communication Style* | Empathetic, Informal, Mixed, Humorous, Direct, Formal |
| *Interests and Hobbies* | Video Games, Painting, Soccer, Reading, Cooking, Traveling, Sports |
| *Educational Background* | High School, Graduate Degree, Self-taught, Bachelor |
| *Cultural Background* | Middle Eastern, Western, Eastern, Latin American, African |
| *Language Proficiency* | English, Spanish, Mandarin, English, English and Spanish, French, Spanish, Mandarin |
| *Technology Savviness* | Intermediate, Novice, Expert |
| *Preferred Communication Medium* | Voice, Mixed, Video, Text |
| *Lifestyle* | Sedentary, Active |
| *Values and Beliefs* | Christianity, Environmentalism, Traditional, Humanism, Islam, Atheism |
| *Relationship Status* | Widowed, Divorced, In a relationship, Single, Married |
| *Economic Status* | Low income, High income, Middle income |
| *Health and Wellness* | Health-conscious, Average health, Healthy |
| *Time Availability* | Sporadic, Full-time, Part-time |
| *Problem-solving Approach* | Practical, Creative, Collaborative, Analytical |

The LoRA configuration, shown in the middle section of Figure 2, was designed to optimize the adaptation process. We set the rank dimension (r) to 16 with a matching alpha value of 16, and a dropout rate of 0. The target layers included key transformer components: query, key, and value projections (q_proj, k_proj, v_proj), along with gate projections and embedding layers.

This configuration was implemented through the Unsloth framework (Unsloth, 2024), which enabled 4-bit quantization for improved memory efficiency.

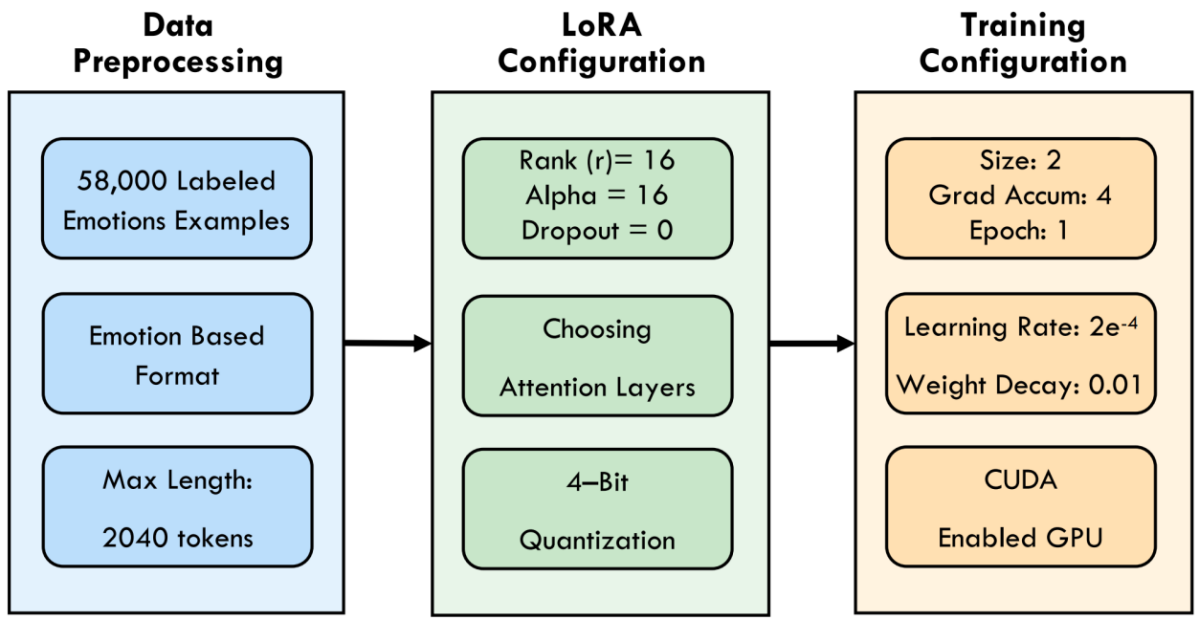

Figure 2. Detailed fine-tuning strategy and implementation architecture

The training configuration, detailed in the bottom section of Figure 2, was optimized for both performance and resource utilization. We employed a batch size of 2 with gradient accumulation steps of 4, effectively creating larger virtual batches while managing memory constraints. The learning rate was set to $2.e^{-4}$ with a weight decay of 0.01, complemented by 5 warmup steps. The entire process was accelerated using CUDA-enabled GPU hardware with specific memory optimizations.

## 2.5. Experimental Setup

Our experimental framework was implemented in a high-performance computing environment, utilizing GPU-accelerated hardware for both fine-tuning and inference tasks. The software infrastructure combined PyTorch for deep learning operations, Hugging Face's Transformers library for model handling, and custom Python scripts for data processing and analysis. The evaluation system was designed to comprehensively assess model responses through multiple metrics. We measured distance accuracy within the specified range of 1,411-1,441 miles around the true value of 1,426 miles, analyzing the optimal subset size required for maximum accuracy, and examining the statistical distribution of response patterns across different configurations.

The selection of the distance estimation task between Fargo, ND and Seattle, WA as our evaluation prompt was deliberate. This question provides an objectively verifiable answer, requires complex geographical reasoning, and maintains neutrality regarding emotional or social factors. These characteristics make it an ideal benchmark for testing collective intelligence capabilities. To evaluate these prompt variations systematically, we developed a comprehensive evaluation framework. The framework has assessed responses based on numerical accuracy within the acceptable range, consistency across multiple runs with similar configurations, and improvements in accuracy through response aggregation. Through this structured approach, we could systematically analyze the impact of emotional integration and social attributes on the model's collective intelligence capabilities.

Table 2. Types, description and components of prompt variation

| Prompt Type | Description | Components |
|---|---|---|
| *Full Context* | Complete prompt with all elements | Role system message, Persona attributes, Emotional state, Core question |
| *Emotional-Only* | Emotional context without attributes | Emotional state, Core question |
| *Attributes-Only* | Social attributes without emotion | Role system message, Persona attributes, Core question |
| *Base* | Question without additional context | Core question only |

## 2.6. Execution Steps

Our experimental execution followed a structured three-phase approach to systematically evaluate the impact of emotional integration on collective intelligence capabilities. The design ensured consistent testing conditions across all phases while addressing variations in prompt composition. During the baseline evaluation phase, we first assessed the original DarkIdol-Llama-3.1-8B model's performance using all four prompt variations outlined in Table 2. It is important to note that Emotional-Only and Base prompts excluded persona attributes, focusing solely on emotional context or the core question, respectively. For each prompt type, 15,064 unique combinations of social attributes were tested where applicable (i.e., in Full Context and Attributes-Only prompts). This baseline evaluation captured response patterns, accuracy rates, and optimal subset sizes for aggregation, establishing control metrics before any emotional fine-tuning.

Next, the sequential testing phase systematically evaluated the impact of each prompt variation. Each of the 15,064 persona configurations was processed sequentially through applicable prompt types. Emotional-Only and Base prompts were excluded from persona-based testing, focusing instead on the interplay between emotional cues and the core question. Controlled batching procedures were employed to optimize computational resource usage, ensuring reproducibility and efficiency. Data collection encompassed both individual response accuracy and aggregate performance metrics for each prompt type.

Finally, the post-fine-tuning evaluation phase replicated the testing protocol with the emotionally enhanced model. After integrating the GoEmotions dataset through the LoRA fine-tuning procedure, we conducted comprehensive testing using the same prompt variations and persona configurations. Emotional-Only and Base prompts once again have excluded persona attributes, maintaining their original contextual focus. By replicating the baseline evaluation under identical conditions, we achieved a direct comparison of performance metrics. This allowed us to quantify the specific impacts of emotional integration on distance estimation accuracy and collective wisdom effects.

## 3. Results and Discussions

This section presents the results of our investigation into the integration of emotional contexts in LLMs and their impact on collective intelligence. We analyze the model's performance across different prompting strategies, comparing pre- and post-fine-tuning configurations to evaluate the influence of emotional diversity and social attributes on decision-making accuracy and efficiency. The findings are discussed in terms of their implications for optimizing emotional integration in LLMs, highlighting key trade-offs between accuracy, efficiency, and contextual complexity.

### 3.1. Quantitative Analysis - Prompting

Our investigation into emotional integration's impact on LLM collective intelligence revealed fascinating patterns in how different types of contextual information influence model performance. Baseline measurements in Table 3 showed a clear hierarchy in prompting effectiveness, with the attribute-only approach achieving the highest accuracy at 92.66% with an optimal subset size of 1,076 responses. This indicates that social attributes provide particularly strong guidance for predictions.

The concept of optimal subset size reflects the smallest number of responses required to achieve near-maximum accuracy without excessive increases in size that yield diminishing returns. Larger subsets may slightly improve accuracy but introduce inefficiencies, such as higher computational costs and redundancy. By identifying this balance, we ensured both accuracy and efficiency in our results.

Emotion-only and combined approaches achieved similar accuracies around 75%, with optimal subset sizes of 2,152 and 538 responses, respectively. The larger subset size for emotion-only prompts highlights the variability in emotional cues, while the combined approach benefits from the synergy between emotional and social contexts, achieving comparable results with fewer responses. This understanding of optimal subset size underscores its critical role in designing efficient and scalable systems.

**Table 3:** Performance Comparison of Different Prompting Strategies Before Fine-tuning

| Data | Optimal Subset Size | Accuracy (%) | Size |
|---|---|---|---|
| *Attributes* | 1076 | 92.66 | 15064 roles |
| *Emotions* | 2152 | 75.05 | 15064 roles |
| *Both* | 538 | 75.46 | 15064 roles |
| *Only Prompt* | 3228 | 15.55 | 15064 roles |

These numerical shifts become sharply focused when examining our graphical analyses. Figure 3, showing the base prompt configuration, establishes our control scenario with its modest 15.55% accuracy. Figure 4 illustrates the remarkable stability and effectiveness of attribute-based prompting, displaying a characteristic sharp peak at optimal performance.

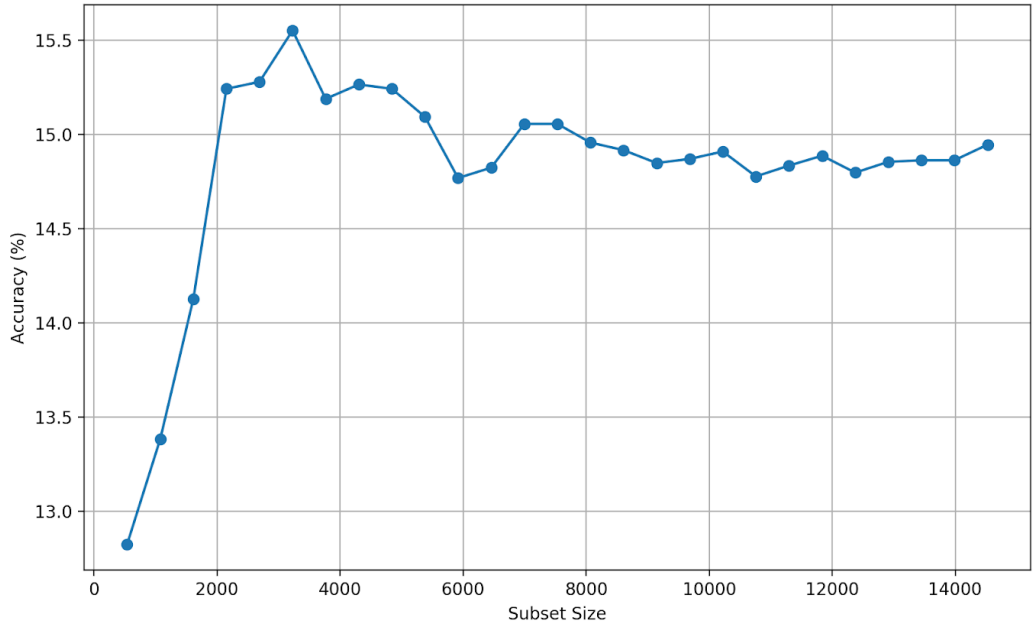

Figure 3. Accuracy vs subset size for base prompt configuration (no context)

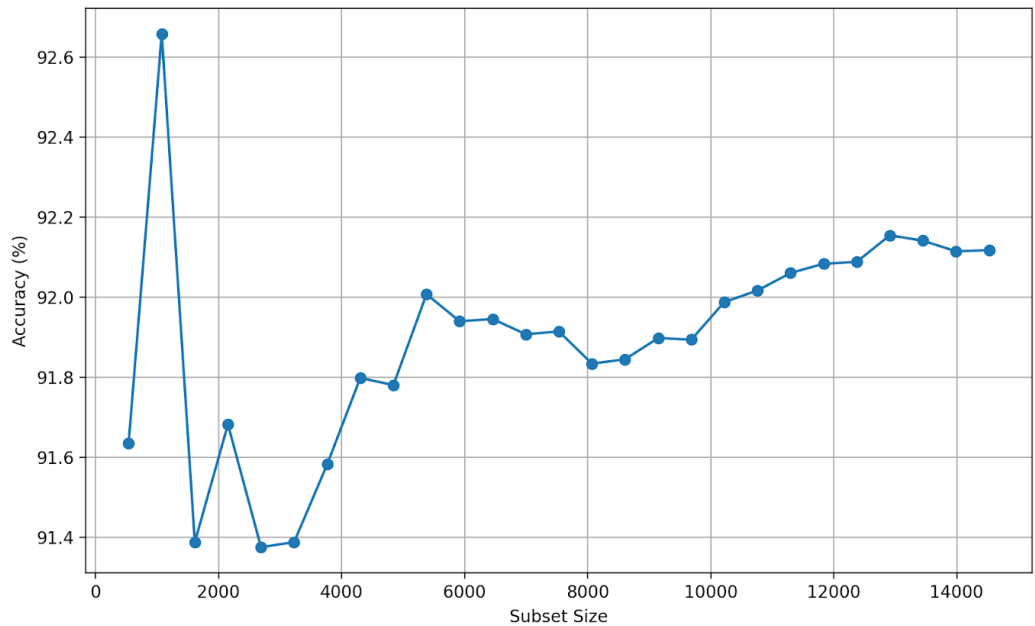

Figure 4. Accuracy vs subset size for attribute-only prompting

The most revealing insights emerge from our emotion-based analyses. Figure 5 shows the pre-fine-tuning performance of emotion-only prompting, demonstrating strong initial accuracy around 75.05%. Figure 6 captures the balanced performance of combined prompting approaches pre-fine-tuning, showing how the interaction between emotional and attribute-based information led to synergistic results with a smaller optimal subset size compared to emotion-only prompts.

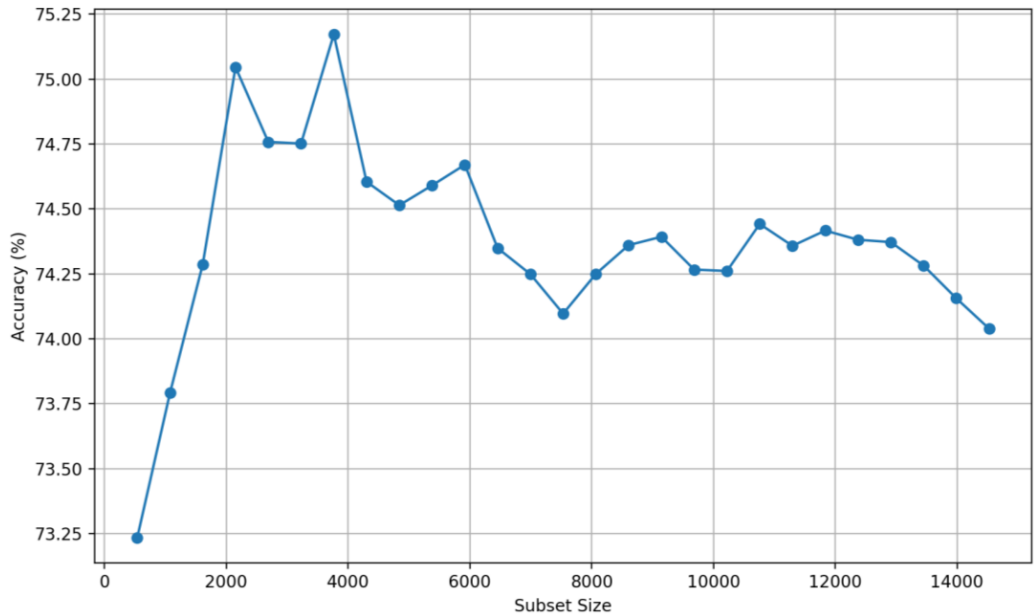

Figure 5. Comparison of accuracy vs subset size for emotion-only prompting: pre-fine-tuning analysis

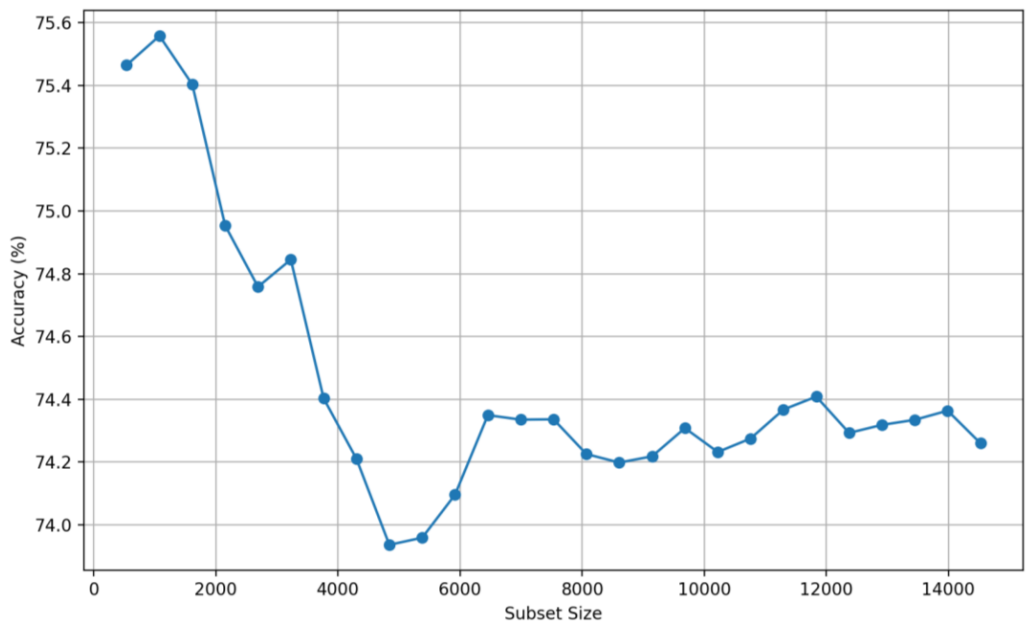

Figure 6. Comparison of accuracy vs subset size for combined (emotions and attributes) prompting: pre-fine-tuning analysis

This comprehensive analysis demonstrates that contextual information significantly influences model performance before fine-tuning, with the attribute-only and combined prompting approaches offering unique benefits. The attribute-only approach provides high raw accuracy, while combined prompting achieves a balance of contextual efficiency and accuracy.

### 3.2. Quantitative Analysis - Fine-Tuning

The transformation brought about by emotional fine-tuning becomes evident in Table 4. The emotion-only approach, while showing decreased accuracy at 36.99%, demonstrated interesting efficiency improvements with a reduced optimal subset size of 538 responses. The combined approach underwent an even more dramatic shift, with accuracy settling at 31.68% but requiring a much larger optimal subset of 4,842 responses.

Table 4. Performance Metrics After Emotional Fine-tuning (focusing on the results of emotion-only and combined approaches post-fine-tuning)

| Data | Optimal Subset Size | Accuracy (%) | Size |
|---|---|---|---|
| *Emotions* | 538 | 36.99 | 15064 roles |
| *Both* | 4842 | 31.68 | 15064 roles |

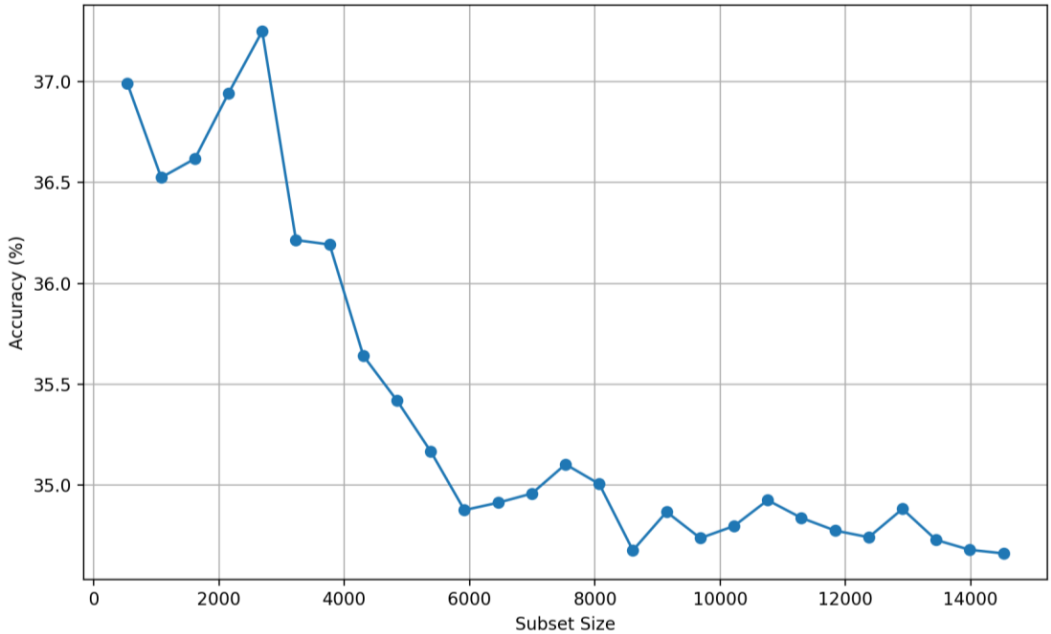

Figure 7. Comparison of accuracy vs subset size for emotion-only prompting: post fine-tuning analysis

The emotional fine-tuning process fundamentally altered how the model processes contextual information. While overall numerical accuracy decreased, the reduced optimal subset size for emotion-only prompts indicates increased efficiency in handling emotional contexts, possibly reflecting a refinement in how the model aggregates emotionally influenced responses. In contrast, the combined approach required a much larger subset size post-fine-tuning, suggesting that integrating emotional and attribute-based information introduced complexities that required more extensive aggregation to achieve stability. Figure 7 captures the changes in emotion-only prompting post-fine-tuning, showing a trade-off between raw accuracy and efficiency gains. Figure 8 highlights the post-fine-tuning evolution of combined approaches, revealing how the interaction between emotional and social attributes became more complex, demanding larger response sets for optimal performance.

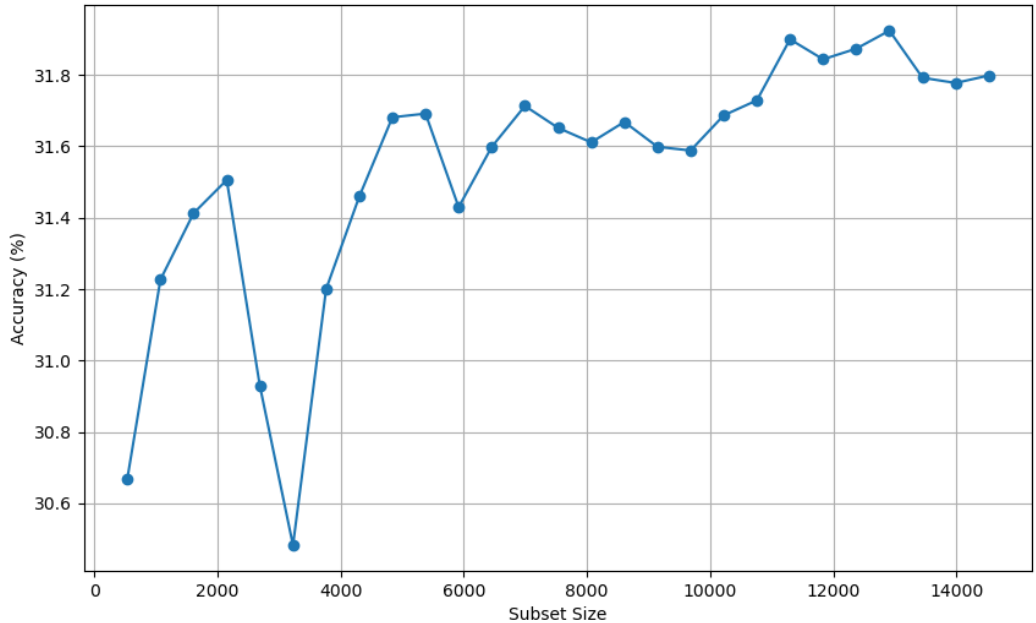

**Figure 8:** Comparison of Accuracy vs Subset Size for Combined (Emotions and Attributes) Prompting: Post Fine-tuning Analysis

These results suggest that emotional fine-tuning introduces trade-offs in artificial collective intelligence systems. While raw accuracy may decrease, the transformation in response patterns and subset utilization indicates a more nuanced approach to information processing, particularly in managing emotional contexts. This nuanced processing capability may open up new avenues for designing emotionally-aware language models that balance predictive accuracy with deeper contextual understanding.

### 3.3. Discussion

The integration of emotional contexts into LLMs reveals complex dynamics in artificial collective intelligence. Our research demonstrates both the expected and surprising effects of emotional fine-tuning on the model's predictive capabilities, while also highlighting important considerations for future development in this field. The impact of emotional integration is manifested in several significant ways. Before fine-tuning, we observed that social attributes provided the strongest framework for accurate predictions, achieving 92.66% accuracy with a relatively small optimal subset size. This suggests that concrete social contexts offer reliable anchoring points for the model's decision-making processes. However, the transformation after emotional fine-tuning tells a more nuanced story. While overall accuracy decreased, we witnessed a fundamental shift in how the model processes information, particularly in its ability to utilize smaller groups more effectively for emotion-based predictions.

When compared with our baseline model, which achieved only 15.55% accuracy with a large optimal subset size of 3,228 responses, both pre-and post-fine-tuning configurations demonstrated substantial improvements. The baseline's poor performance underscores the critical role of contextual information in guiding model predictions. Interestingly, even after the accuracy decreased following emotional fine-tuning, the model's performance remained significantly above baseline, suggesting that emotional integration, while potentially disruptive to pure numerical accuracy, maintains valuable predictive capabilities.

The relationship between emotions and social attributes proved particularly fascinating. Before fine-tuning, their combination achieved comparable accuracy to emotion-only prompting (75.46% vs 75.05%) but required fewer responses for optimal performance. After fine-tuning, this dynamic shifted dramatically, with the combined approach requiring a much larger subset size (4,842 responses) to achieve its best performance. This suggests that emotional fine-tuning creates more complex interactions between emotional and social contexts, requiring larger sample sizes to effectively aggregate these multiple dimensions of information.

However, our research faced several important limitations. First, the distance estimation task, while providing a clear benchmark for accuracy, may not fully capture the nuanced ways emotional integration affects model performance in other types of predictions. Second, our fine-tuning process, focused solely on the GoEmotions dataset, might not represent the full spectrum of emotional expressions and their interactions with decision-making processes. The dramatic decrease in numerical accuracy post-fine-tuning, while revealing interesting patterns in information processing, also suggests potential limitations in our current approach to balancing emotional awareness with predictive precision.

The trade-off between accuracy and emotional integration raises important questions about the nature of artificial collective intelligence. Just as human groups might sacrifice some analytical precision for better emotional understanding (Zhu et al., 2023), our model appears to undergo a similar transformation. This suggests that developing more sophisticated collective intelligence systems might require embracing such trade-offs rather than trying to optimize for accuracy alone.

These findings point toward several promising directions for future research, including investigating methods to better balance emotional awareness with predictive accuracy, exploring how different types of emotional contexts affect various prediction tasks, and developing more sophisticated approaches to integrating multiple dimensions of contextual information in language models.

## 4. Conclusion

The integration of emotional contexts into LLMs to enhance collective intelligence has revealed significant insights into how artificial systems can simulate collective intelligence. Our research, utilizing the DarkIdol-Llama-3.1-8B model and the GoEmotions dataset, demonstrates both the potential and challenges of incorporating emotional diversity into artificial collective intelligence systems.

Our key findings reveal intriguing patterns in how different types of contextual information influence model performance. Pre-fine-tuning results showed that social attributes alone achieved the highest accuracy (92.66%), while emotional and combined approaches maintained a strong performance at around 75%. The transformation after emotional fine-tuning, though showing decreased numerical accuracy, demonstrated interesting shifts in how the model processes information, particularly in optimal subset sizes for peak performance. Notably, the emotion-only approach post fine-tuning achieved 36.99% accuracy with a smaller optimal subset size of 538 responses, suggesting more efficient emotional information processing. The combined approach, requiring 4,842 responses for 31.68% accuracy, indicates a more complex integration of emotional and social contexts after fine-tuning.

These results suggest that while emotional integration may initially reduce numerical accuracy, it fundamentally transforms how the model processes and combines different types of contextual information. This trade-off between emotional awareness and predictive precision mirrors human group dynamics, where emotional intelligence sometimes operates at the expense of pure analytical accuracy.

Future research could explore alternative fine-tuning techniques to better balance emotional awareness with predictive accuracy and investigate the impact of different emotional datasets beyond GoEmotions. Developing hybrid approaches that maintain high attribute-based accuracy while incorporating emotional intelligence presents an exciting direction. Additionally, examining how emotional integration affects performance across different types of prediction tasks could provide valuable insights into the broader applicability of these findings. The optimization of multiple contextual dimensions in language models and the relationship between subset size and emotional processing in various task contexts also merit further investigation.

Our findings contribute to the growing understanding of how emotional diversity influences artificial collective intelligence, suggesting new approaches for developing more sophisticated and human-like crowd wisdom simulations in AI systems. This research opens new avenues for creating AI systems that can better balance analytical precision with emotional understanding, potentially leading to more nuanced and effective collective intelligence applications.

**Funding**

No funds, grants, or other support was received.

**Competing Interest Declaration**

The authors have no relevant financial or non-financial interests to disclose.

**Data Availability**

All data that is produced and analyzed in the manuscript is readily available and presented in the manuscript.

**Declaration of Generative AI and AI-Assisted Technologies**

During the preparation of this manuscript, the authors used ChatGPT, based on the GPT-4 model, to improve the flow of the text, correct grammatical errors, and enhance the clarity of the writing. The language model was not used to generate content, citations, or verify facts. After using this tool, the authors thoroughly reviewed and edited the content to ensure accuracy, validity, and originality, and take full responsibility for the final version of the manuscript.

**CRediT author statement**

**Likith Kadiyala**: Conceptualization, Methodology, Software, Validation, Formal analysis, Investigation, Data Curation, Writing - Original Draft, and Visualization. **Ramteja Sajja**: Validation, Methodology, Writing - Review & Editing. **Yusuf Sermet**: Conceptualization, Methodology, Writing - Review & Editing, Investigation, Project administration, and Validation. **Ibrahim Demir**: Conceptualization, Methodology, Writing - Review & Editing, Supervision, Funding acquisition, and Resources.